# An Land Cover Fuzzy Logic Classification By Maximumlikelihood

T.Sarath[#1], G.Nagalakshmi[*2]

[#1]M.Tech, Department Of CSE, Sistk , Puttur. A.P., India

[*2]Professor & Hod, Department Of CSE, Sistk, Puttur, A.P.,India

*Abstract*— In present days remote sensing is most used application in many sectors. This remote sensing uses different images like multispectral, hyper spectral or ultra spectral. The remote sensing image classification is one of the significant method to classify image. In this state we classify the maximum likelihood classification with fuzzy logic. In this we experimenting fuzzy logic like spatial, spectral texture methods in that different sub methods to be used for image classification.

## I. INTRODUCTION

The success of any GIS [10, 11] application depends on the quality of the geographical data used. GIS [15] means "Geographic Information System". In general it defined as computer assisted systems for the capture, storage, retrieval, analysis and display of spatial/spectral data. This will collect the high-quality geographical data for input. To study this we take an remote sensing.

### A. Hyper spectral Remote Sensing

The hyper spectral remote sensing [1, 8, 2, 17] is an advanced tool that provides high spatial/spectral resolution data from a distance.

The most powerful tools used in the field of remote sensing are Hyper spectral imaging (HSI) and Multispectral Imaging (MSI)

Since the mid 1950's some airborne sensors have recoded spectral information [8] on the Earth surface in the wavelength region extending from 400 to 2500 nm. Starting from the early 1970's, [9] a large number of space borne multispectral sensors have been launched, on board the LANDSAT, SPOT and Indian Remote Sensing (IRS) series of satellites.

Hyper spectral image like other spectral image which collects information from across the electromagnetic spectrum. Such as the human eye sees visible light in three bands (red, green, and blue), spectral imaging divides the spectrum into many more bands. This technique of dividing images into bands which is extended beyond the visible.

Engineers build sensors and processing systems to provide such capability for application in agriculture, mineralogy, physics, and surveillance. Hyper spectral sensors look at objects using a vast portion of the electromagnetic spectrum. Certain objects leave unique 'fingerprints' across the electromagnetic spectrum. These 'fingerprints' are known as spectral signatures and enable identification of the materials that make up a scanned object. For example, a spectral signature for oil helps mineralogists find new oil fields.

Hyper spectral imaging is part of a class of techniques commonly referred to as spectral imaging or spectral analysis. Hyper spectral imaging is related to multispectral imaging. The distinction between hyper and multi-spectral is sometimes based on an arbitrary "number of bands" or on the type of measurement, depending on what is appropriate to the purpose.

Multispectral image deals with several images at discrete and somewhat narrow bands. Being "discrete and somewhat narrow" is what distinguishes multispectral in the visible from color photography. A multispectral sensor may have many bands covering the spectrum from the visible to the long wave infrared. Multispectral images do not produce the "spectrum" of an object. Land sat is an excellent example of multispectral imaging.

Hyper spectral deals with imaging narrow spectral bands over a continuous spectral range, and produce the spectra of all pixels in the scene. So a sensor with only 20 bands can also be hyper spectral when it covers the range from 500 to 700 nm with 20 bands each 10 nm wide. (While a sensor with 20 discrete bands covering the VIS, NIR, SWIR, MWIR, and LWIR would be considered multispectral.)

'Ultra spectral' could be reserved for interferometer type imaging sensors with a very fine spectral resolution. These sensors often have (but not necessarily) a low spatial resolution of several pixels only, a restriction imposed by the high data rate.

In this we do the hyper spectral remote sensing classification, where image classification is a process of sorting pixels in to individual classes, based on pixel values. This classification is used to assign corresponding levels with respect to groups. This classification is mostly used as extraction techniques in digital remote sensing. Most of the digital image analysis is very nice to have a good image to show a magnitude of colours contains various features of the underlying terrain, but it is useless if you don't know what the colours mean.

There are two main classification methods are Supervised Classification and Unsupervised Classification. The unsupervised classification is the identification of natural groups. The supervised classification is the process of sampling the known identity to classify and unclassified pixels to one of several informational classes.

## II. MAXIMUM LIKELIHOOD CLASSIFICATION

Maximum likelihood classification [18, 19] (MLC Pixel based) Maximum likelihood decision rule is based on Gaussian estimate of the probability density function of each class (Pedroni, 2003). Maximum likelihood classifier evaluates both the variance and covariance of the spectral response patterns in classifying an unknown pixel. It assumes the distribution of the cloud of points forming the category training data to be normally distributed. Under this assumption, distribution of response pattern can be described by mean vector and the covariance matrix. From the given parameters the statistical probability of a given pixel value can be computed. By computing the probability of the pixel value, an undefined pixel can be classified. After evaluating the probability the pixel would be assigned to the one with highest probability value.

One of the drawbacks in maximum likelihood classifier is large number of computation required to classify each pixel. This is true when large number of spectral classes must be differentiated. Suppose there is a sample $X_1, X_2,..., X_n$ of n independent and identically distributed observations, coming from a distribution with an unknown probability density function $f_0(\cdot)$. It is however surmised that the function f0 belongs to a certain family of distributions $\{f(\cdot|\theta), \theta \in \Theta\}$ (where θ is a vector of parameters for this family), called the parametric model, so that $f_0 = f(\cdot|\theta 0)$. The value θ is unknown and is referred to as the true value of the parameter. It is desirable to find an estimator Ө which would be as close to the true value θ0 as possible. Both the observed variables xi and the parameter θ can be vectors.

To use the method of maximum likelihood, one first specifies the joint density function for all observations. For an independent and identically distributed sample, this joint density function is

$$F(x_1, x_2, …,x_n/\Theta)=f(x_1/\Theta)\text{x}f(x_2/\Theta)\text{x}….\text{x}f(x_n/\Theta)$$

Now we look at this function from a different perspective by considering the observed values $x_1, x_2, ... , x_n$ to be fixed "parameters" of this function, whereas θ will be the function's variable and allowed to vary freely; this function will be called the likelihood:

$$£(\theta/x_1,……x_n) = f(x_1,x_2,….x_n/\theta) = \prod_{i=1}^{n} f(x_i/\theta)$$

In the exposition above, it is assumed that the data are independent and identically distributed. The method can be applied however to a broader setting, as long as it is possible to write the joint density function $f(x_1, … , x_n | θ)$, and its parameter θ has a finite dimension which does not depend on the sample size n. In a simpler extension, an allowance can be made for data heterogeneity, so that the joint density is equal to $f_1(x_1|θ) \cdot f_2(x_2|θ) \cdot \cdots \cdot f_n(x_n | θ)$. Put another way, we are now assuming that each observation xi comes from a random variable that has its own distribution function fi. In the more complicated case of time series models, the independence assumption may have to be dropped as well.

A maximum likelihood estimator coincides with the most probable Bayesian estimator given a uniform prior distribution on the parameters. Indeed, the maximum a posteriori estimate is the parameter θ that maximizes the probability of θ given the data, given by Baye's theorem:

$P(\Theta/x_1,x_2,….,x_n)=f(x_1,x_2,….,x_n/\Theta)P(\Theta)/P(x_1, x_2,…, x_n)$

Where $P(\Theta)$ is the prior distribution for the parameter θ and where $P(x_1,x_2,….x_n)$ is the probability of the data averaged over all parameters. Since the denominator is independent of θ, the Bayesian estimator is obtained by maximizing $f(x_1,x_2,….,x_n/\Theta)P(\Theta)$ with respect to θ. If we further assume that the prior P (Ө) is a uniform distribution, the Bayesian estimator obtained by maximizing the likelihood function $f(x_1,x_2,….,x_n/\Theta)$. Thus the Bayesian estimator coincides with the maximum-likelihood estimator for a uniform prior distribution P (Ө).

Maximum likelihood classification assumes that the statistics for each class in each band are normally distributed and calculates the probability that a given pixel belongs to a specific class. Unless you select a probability threshold, all pixels are classified. Each pixel is assigned to the class that has the highest probability (that is, the maximum likelihood). If the highest probability is smaller than a threshold you specify, the pixel remains unclassified.

ENVI implements maximum likelihood classification by calculating the following discriminate functions for each pixel φ in the image (Richards, 1999):

$$g_i(x) = 1np(w_i) - \frac{1}{2}1np\left|\sum i\right| - \frac{1}{2}(x-n_i)^2 \sum_i^{-1}(x-m_i)$$

Where:

i = class
x = n-dimensional data (where n is the number of bands)
p ($\omega i$) = probability that class $\omega i$ occurs in the image and is assumed the same for all classes
|Σi| = determinant of the covariance matrix of the data in class $\omega i$
Σi-1 = its inverse matrix
$M_i$ = mean vector.

## III. METHODOLGY

*A. Fuzzy logic*

Traditional rule-based classification is based on strict binary rules, for example: objects meeting the rules for "tree" are classified as "tree," objects meeting the rules for "urban" are classified as "urban," and objects meeting neither rule remain unclassified. Fuzzy logic [20] is an important element in ENVI Feature Extraction rule-based classification. Rather than classifying an object as fully "true" or "false" (as in binary rules), fuzzy logic uses a membership function to represent the degree than an object belongs to a feature type. Information extraction from remote sensing data is limited by noisy sensor measurements with limited spectral and spatial resolution, signal degradation from image pre-processing, and imprecise transitions between land-use classes. Most remote sensing images contain mixed pixels that belong to one or more classes. Fuzzy logic helps alleviate this problem by simulating uncertainty or partial information that is consistent with human reasoning. The output of each fuzzy rule is a confidence map, where values represent the degree that an object belongs to the feature type defined by this rule. In classification, the object is assigned to the feature type that has the maximum confidence value. With rule-based classification, you can control the degree of fuzzy logic of each condition when you build rules.

*B. Fuzzy Maximum Likelihood Classification*

The fuzzy set theory [22] can be extended to the maximum likelihood algorithm [21] to measure the membership grade of the pixels. This extension is used by Wang (1990) and Maselli et al (1995).

Based on probability theory, if an event A is a precisely defined set of elements in the universe of discourse $\varphi$, the probability density function of A denoted by p(A) can be expressed by

$$P(A) = \int_\varphi H_A(S) \qquad 1$$

Where S is an elements in $\varphi$, $H_A$ is a membership function, $H_A(S) = 0\ or\ 1$. In the case of image classification, event A is the cluster or class and S is a vector of feature values associated with a specific pixel (The membership grade is 1) or not (i.e. membership grade is zero).

If A is regarded as a fuzzy set which means that set A is a fuzzy subset in $\varphi$, a probability measure of A becomes:

$$P(A) = \int_\varphi \mu_A(S) \qquad 2$$

The term $\mu_A$ is the fuzzy membership function as defined in equation2 is an extension and generalization of equation1. Even partial membership value of observations in A can provide a contribution to the total probability P(A).

The mean and variance of fuzzy set A relative to a probability measure can similarly be quantified as

$$V_A = \frac{1}{P(A)} \int_\varphi S \mu_A(S) \qquad 3$$

And

$$\sigma_A^2 = \int_\varphi (S - V_A)^2 \mu_A(S) \qquad 4$$

Equations 3&4 determine the fuzzy mean and fuzzy variance, respectively both of which are derived for the continuous case. In practice, the discrete fuzzy mean V and fuzzy covariance matrix F for class I can be expressed as

$$V_i = \frac{\sum_i \mu_i(X_i) X_i}{\sum_i \mu_i(X_i)} \qquad 5$$

And

$$F_i = \frac{\sum_i \mu_2(X_i)(X_i - V_i)(X_t - V_i)^T}{\sum_i \mu_i(X_i)} \qquad 6$$

Where $X_i$ denotes the feature vector for pixel j. if the exponent m in equation6 is set to 1 then it becomes equivalent, which is used in the optimum clustering algorithm. It can be preferred that when the value of $\mu_i(X_i)$ becomes either 0 or 1.

A fuzzy set is characterized by its membership function Wang (1990) defines the membership grade for each land cover class based on the maximum likelihood classification algorithm with fuzzy mean and fuzzy covariance matrix as shown in equation 5&6 as follows

$$\mu_k(X_i) = \frac{P_K(X_i)}{\sum_i P_i(X_i)} \qquad 7$$

Where k is the land cover class and probability $P_i(X_i)$ denotes the class conditional probability for class I given the observation $X_i$.

The fuzzy maximum likelihood algorithm calculating membership grades in terms of equation7 is equivalent to normalizing the probabilities of the pixel to all of the information classes. Although such a method is quite straight forward, its validity requires further investigation.

In applications of the fuzzy maximum likelihood approach, it is questionable whether or not mixed pixels should be used to construct the fuzzy mean and covariance matrix as suggested Wang (1990).

### IV. STUDY AREA

These areas results to find the supervised classification on maximum likelihood classification with training sites which include with the reason of interest (ROI). By using this we mainly used to find the different areas to find in an image each reason will represent as a class in which it may take each class as training sites we can represent with polygon or point. By this we can get many bands and to smooth the areas then we will get the maximum likelihood classification image as shown in fig2. This image we will get in the ENVI zoom. By using the maximum likelihood classification image we will conduct the feature extraction to that image and select the band to that select scale level and merge level to find the refinement in thresholding advanced state is selected for that state it has spectral, spatial, texture state in that we have to choose the creating rules and in that we have select the add attribute to rule in that we can select each state have different methods. In the methods at texture we selected tx_mean in spectral we selected avgband and in spatial majaxilan and we have to select the fuzzy tolerance and to set the function type may be s_type or linear and to find the vector level to be leveled so that we have to smooth the level as the respected output will be the fuzzy maximum likelihood classification image as shown in fig3. The performance of the fuzzy maximum likelihood classification is as shown in TABLEII

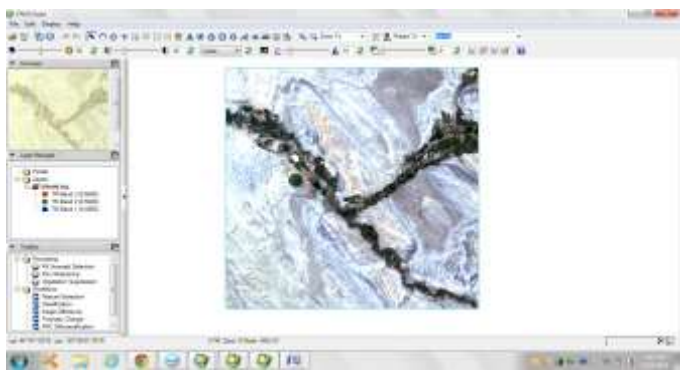

Fig. 1 Input Image

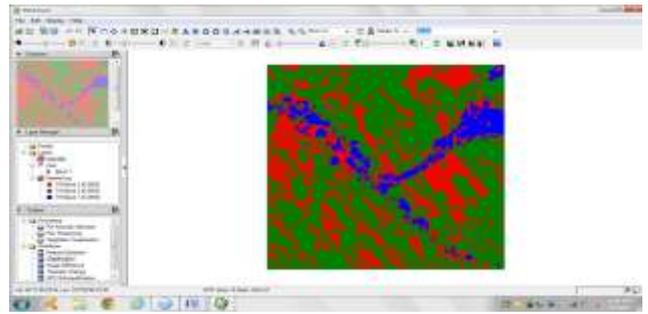

Fig. 2 Maximum Likelihood Classification Image

TABLE I

SHOWS THE BANDWIDTHS OF MAXIMUM LIKELIHOOD CLASSIFICATION

| S. No. | No. of Bands | Value |
|---|---|---|
| 1 | Band 1 | 0.4850 |
| 2 | Band 2 | 0.5600 |
| 3 | Band 3 | 0.6600 |

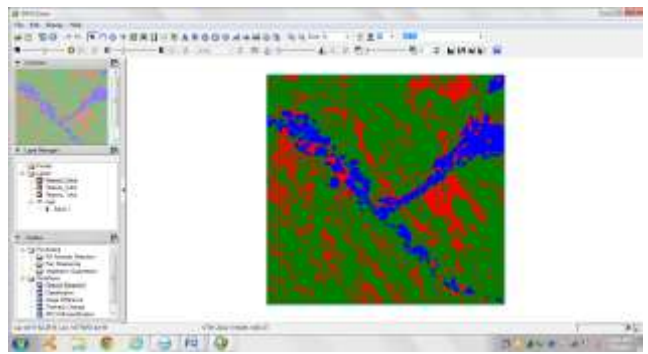

Fig. 3 Fuzzy Maximum Likelihood Classification Image

### V. PERFORMANCE

File Name: maxi
Segment Scale Level:    50.0
Merge Level:            0.0
Refine:            1.00000 to 3.00000
Attributes Computed:
Spatial
Spectral
Texture
Classification: Rule-Based
Rule Set:
1. (1.000): If tx_mean [0.7242, 2.8601], then object belongs to "Feature_1".
2. (1.000): If avgband_1 < 2.0131, then object belongs to "Feature_2".
3. (1.000): If majaxislen < 2047.6970, then object belongs to "Feature_3".

Export Options:
  Vector Output Directory: C:\Users\folder\AppData\Local\Temp\

Feature Info:
Feature_1 Type: Polygon
Feature_2 Type: Polygon

Feature_3 Type: Polygon
Smoothing: Threshold of 1

TABLE II

PERFORMANCE OF THE FUZZY MAXIMUMLIKELIHOOD CLASSIFICATION

| Feature Name | Feature Count | Total Area | Mean Area | Min Area | Max Area |
|---|---|---|---|---|---|
| Feature_1 | 51 | 188128800 | 3688800 | 2700 | 185912100 |
| Feature_2 | 118 | 27000450 | 228817.37 | 1800 | 5563350 |
| Feature_3 | 82 | 12077550 | 147287.2 | 1800 | 2325600 |

## VI. CONCLUSION

In this paper we conduct the fuzzy logic by rule based classification for spatial, spectral, texture methods we only classified tx_mean, avgbands, and majaxislen and for that feature count, total area, mean area, min area, max area We can conduct all the stages in the rule based method which will show the feature extraction and we can state different methods of spectral, spatial, texture. In this we only conducted feature extraction on maximum likelihood but we can conducted on different methods of supervised classification and unsupervised classification.